\documentclass[runningheads]{llncs}
\usepackage{graphicx}
\usepackage{graphics}
\usepackage{subfigure}
\usepackage{amsmath}
\usepackage{algorithm}
\usepackage{algorithmic}

\begin{document}

\title{Saliency Detection via Bidirectional Absorbing Markov Chain}
%
%
\author{Fengling Jiang\inst{1,2,3}
 \and Bin Kong\inst{1,4} \and Ahsan Adeel \inst{5} \and Yun Xiao \inst {6} \and Amir Hussain\inst{5} }
%
\authorrunning{F. Jiang et al.}
%
\institute{Institute of Intelligent Machines, Chinese Academy of Sciences, Hefei 230031, China
\and University of Science and Technology of China, Hefei 230026, China
\and Hefei Normal University, Hefei 230061, China
\and Anhui Engineering Laboratory for Intelligent Driving Technology and Application, Hefei 230088, China
\and University of Stirling, Stirling FK9 4LA, UK\\
\and Anhui University, Hefei 230601, China\\
\email{fljiang@mail.ustc.edu.cn, bkong@iim.ac.cn, aad@cs.stir.ac.uk, xiaoyun@ahu.edu.cn, ahu@cs.stir.ac.uk}}
\maketitle              

\begin{abstract}
Traditional saliency detection via Markov chain only consider boundaries nodes. However, in addition to boundaries cues, background prior and foreground prior cues play a complementary role to enhance saliency detection. In this paper, we propose an absorbing Markov chain based saliency detection method considering both boundary information and foreground prior cues. The proposed approach combines both boundaries and foreground prior cues through bidirectional Markov chain. Specifically, the image is first segmented into superpixels and four boundaries nodes (duplicated as virtual nodes) are selected. Subsequently, the absorption time upon transition node's random walk to the absorbing state is calculated to obtain foreground possibility. Simultaneously, foreground prior as the virtual absorbing nodes is used to calculate the absorption time and obtain the background possibility. Finally, two obtained results are fused to obtain the combined saliency map using cost function for further optimization at multi-scale. Experimental results demonstrate the outperformance of our proposed model on 4 benchmark datasets as compared to 17 state-of-the-art methods.
\end{abstract}

\keywords{saliency detection \and Markov chain \and background possibility \and foreground possibility \and bidirectional absorbing.}

\section{Introduction}
Saliency detection aims to effectively highlight the most important pixels in an image. It helps to reduce computing costs and has widely been used in various computer vision applications, such as image segmentation~\cite{arbelaez2011contour}, image retrieval~\cite{yang2015scalable}, object detection~\cite{chang2011fusing,gao2015visual}, object recognition~\cite{ren2014region}, image adaptation~\cite{sun2013image}, and video segmentation~\cite{wang2018saliency,tu2016new}. Saliency detection could be summarized in three methods: bottom-up methods~\cite{riche2012rare,yang2013saliency,zhao2015saliency}, top-down methods~\cite{yang2017top,cholakkal2016backtracking} and mixed methods~\cite{Borji2012Probabilistic,wang2018deep,yan2018unsupervised}. The top-down methods are driven by tasks and could be used in object detection tasks. The authors in~\cite{yang2012top} proposed a top-down method that jointly learns a conditional random field and a discriminative dictionary. Top-down methods could be applied to address complex and special tasks but they lack versatility. The bottom-up methods are driven by data, such as color, light, texture and other basic features. Itti et al~\cite{itti1998model} proposed a saliency method by using these basic features. It could be effectively used for real-time systems. The mixed methods are considered both bottom-up and top-down methods.

In this paper, we focus on the bottom-up methods, the proposed method is based on the properties of Markov model, there are many works based on Markov model, such as~\cite{alkhateeb2011performance,alkhateeb2011offline}. Traditional saliency detection via Markov chain~\cite{jiang2013saliency} is based on Marov model as well, but it only consider boundaries nodes. However, in addition to boundaries cues, background prior and foreground prior cues play a complementary role to enhance saliency detection. We consider four boundaries information and the foreground prior saliency object, using absorbing Markov chain, namely, both boundary absorbing and foreground prior are considered to get background and foreground possibility.
In addition, we further optimize our model by fusing these two possibilities, and exploite multi-scale processing.
Fig.1 demonstrates and compares the results of our proposed method with the traditional saliency detection absorbing Markov chain (MC) method~\cite{jiang2013saliency}, where the outperformance of our method is evident.
\begin{figure}[htb]
  \centering
   \includegraphics[width=1\textwidth]{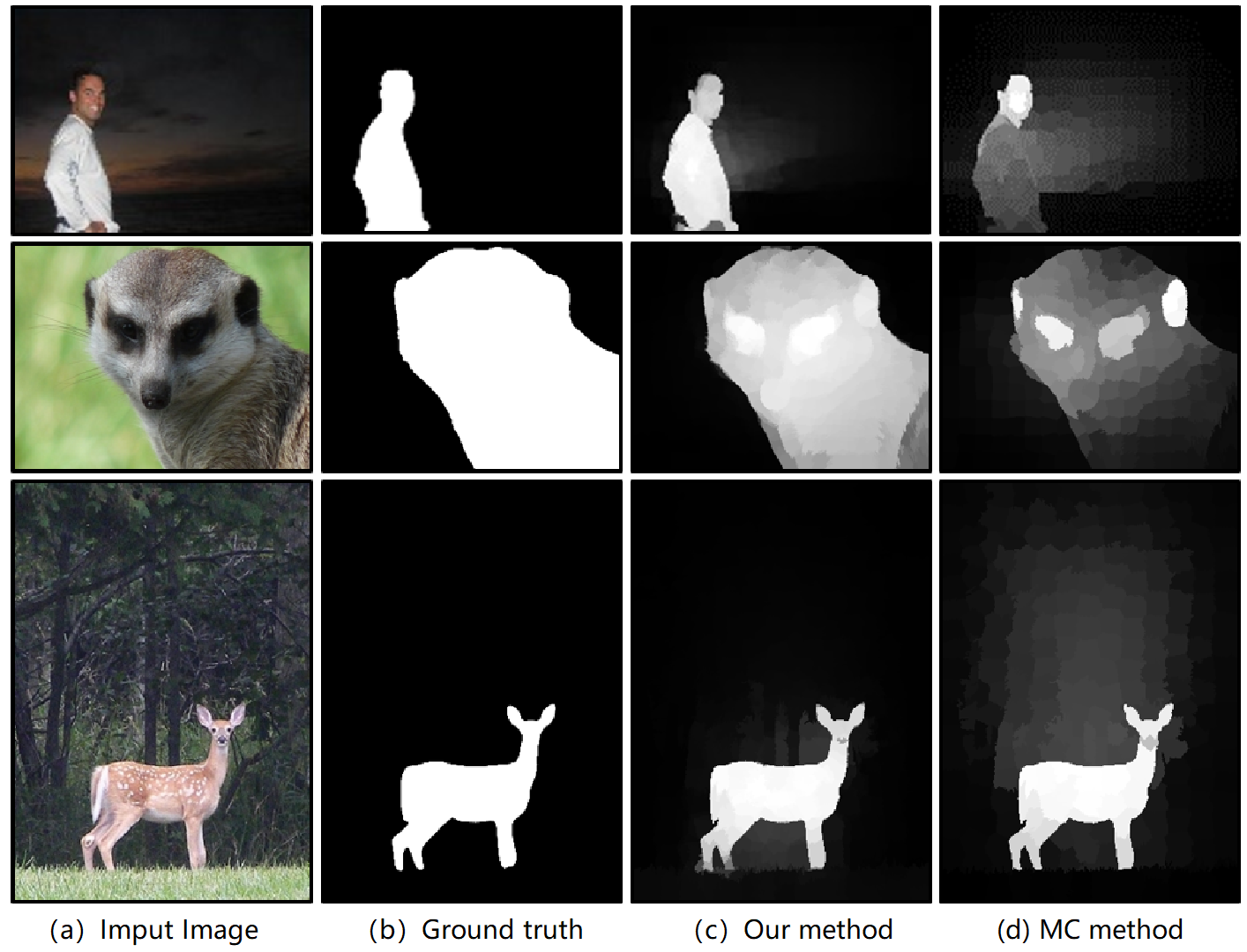}
   \caption{Comparison of the proposed method with the ground truth and MC method.}
\normalsize
\end{figure}

\section{Related works}
There are existing many studies on saliency detection in the past decades according to the resent surveys~\cite{borji2013state,borji2015salient,zhang2018review}. And the proposed model ia based on absorbing Markov chain belonging to bottom-up method, therefore, in this section, we mainly focus on the traditional models and other newly models.

The traditional models which belong to bottom-up methods, have the features of unconscious, fast, data driven, low-level feature driven, which means without any prior knowledge, bottom-up saliency detection can achieve the goal of finding the important regions in a image. The earliest researchers Itti and Koch model~\cite{itti1998model,itti2001computational}, which compute saliency maps though low-level features, such as texture, orientation, intensity, and color contrast. From then on, multitudinous   traditional saliency models have been appeared and acquired outperformances. Some methods based on pixel~\cite{zhai2006visual,shi2013pisa,cheng2013efficient}, Some methods based on superpixel~\cite{wei2012geodesic,margolin2013makes}, and others based on multi-scale~\cite{yan2013hierarchical,jiang2013salient,zhang2018salient}. Jiang et al~\cite{jiang2013saliency} propose saliency detection model by random walk via absorbing Markov Chain where absorbing nodes are duplicated from the four boundaries, and compute absorbing time from the transient nodes to absorbing nodes, obtain the final saliency maps. Considering the
importance of the transition probability matrix, Zhang et al~\cite{zhang2018saliency} based on their aforementioned work propose a learnt transition probability matrix to improve the perfermance. There are some other work based on Markov chain. Zhang et al~\cite{zhang2017two} propose an approach to detection salient objects by exporing both patch-level and object-level cues via absorbing Markov chain. Zhang et al~\cite{zhang2016region} present a data-driven salient region detection model based on absorbing Markov chain via multi-feature. Zhu et al~\cite{zhu2014saliency} integrate boundary connectivity based on geodesic distances into a cost function to get the final optimized saliency map. Li et al~\cite{li2018saliency} propose a saliency optimization scheme by considering both the foreground appearance and background prior. Resent, due to the great developing of deep learning, there are numerous saliency detection models~\cite{zhao2015saliency,wang2018deep,wang2016saliency,lee2018eld} based on deep learning, which obtain outperformance than the traditional methods. However, deep learning based methods need much dates to train, costing much time on computation, which makes the proceeding of saliency detection much complex than before. In this work, based on the traditional method, we introduce bidirectional absorbing Markov chain to this kinds of work to get excellent performance in saliency detection.

\section{Fundamentals of absorbing Markov chain}
In absorbing Markov chain, the transition matrix $P$ is primitive~\cite{charles1997introduction}, by definition, state $i$ is absorbing when $P(i,i)=1$, and $P(i,j)=0$ for all $i \neq j$. If the Markov chain satisfies the following two conditions, it means there is at least one or more absorbing states in the Markov chain. In every state, it is possible to go to an absorbing state in a finite number of steps(not necessarily in one step), then we call it absorbing Markov chain. In an absorbing Markov chain, if a state is not a absorbing state, it is called transient state.

An absorbing chain has $m$ absorbing states and $n$ transient states, the transfer matrix $P$ can be written as:
\begin{equation}
P\rightarrow\left(
  \begin{array}{cc}
    Q&R\\
    0&I\\
  \end{array}
\right),
\end{equation}
where $Q$ is a n-by-n matrix, giving transient probabilities between any transient states, $R$ is a nonzero n-by-m matrix giving these probabilities from transient state to any absorbing state, $0$ is a m-by-n zero matrix and $I$ is the m-by-m identity matrix.

For an absorbing chain $P$, all the transient states can achieve absorbing states in one or more steps, we can write the expected number of times $N(i,j)$ (which means the transient state moves from $i$ state to the $j$ state), its standard form is written as:
\begin{equation}
\label{Eq::transient time}
N=(I-Q)^{-1},
\end{equation}
 namely, the matrix $N$ with invertible matrix, where $n_{ij}$ denotes the average transfer times between transient state $i$ to transient state $j$. Supposing $c=[1,1,\cdot \cdot \cdot,1]_{1\times n}^{N}$, the absorbed time for each transient state can be expressed as:
\begin{equation}
z=N\times c.
\label{Eq::absorb}
\end{equation}
\section{The proposed approach}

To obtain more robust and accurate saliency maps, we propose a method via bidirectional absorbing Markov chain. This section explains the procedure to find the saliency area in an image in two orientations. Simple linear iterative clustering(SLIC) algorithm~\cite{achanta2012slic} has been used to get the superpixels. The pipeline is explained below:
\begin{figure}[htb]
\centering
\includegraphics[width=1\textwidth]{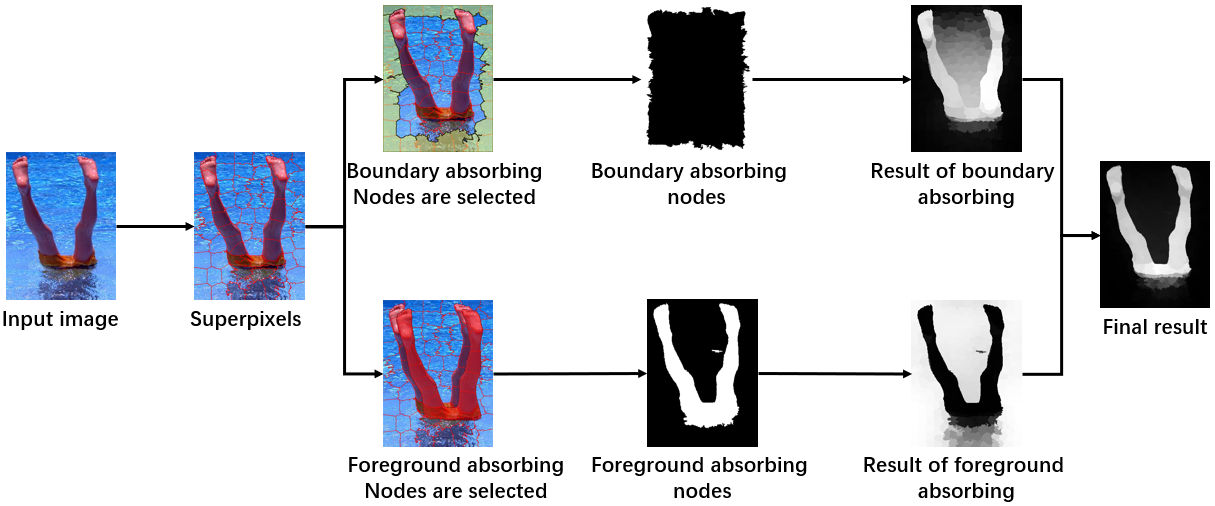}
\caption{The processing of our proposed method}
\label{fig:example}
\end{figure}
\subsection{Graph construction}
\begin{figure}[htb]
\centering
\includegraphics[width=0.6\textwidth]{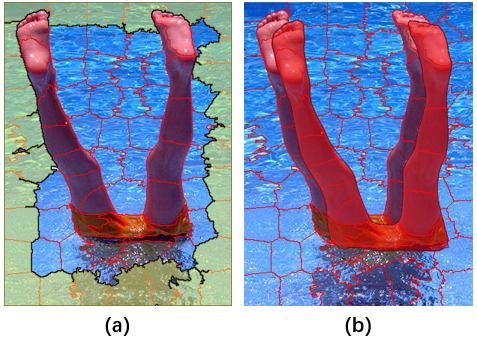}
\caption{The graph construction. (a) The superpixels in the yellow area are the duplicated absorbing nodes. (b) The foreground cues are obtained and regarded as absorbing nodes which are duplicated with red area.}
\label{fig:construction}
\end{figure}

The SLIC algorithm is used to split the image into different pitches of superpixels. Afterwards, two kinds of graphs $G^1(V^1,E^1)$ and $G^2(V^2,E^2)$ are constructed, see Figure~\ref{fig:construction} for detail,
\subsection{Graph construction}

where $G^1$ represents the graph of boundary absorbing process and $G^2$ represents the graph of foreground prior absorbing process. In each of the graphs, $V^1, V^2$ represent the graph nodes and $E^1,E^2$ represent the edges between any nodes in the graphs. For the process of boundary absorbing, superpixels around the four boundaries as the virtual nodes are duplicated. For the process of foreground prior absorbing, superpixels from the regions (calculated by the foreground prior) are duplicated. There are two kinds of nodes in both graphs, transient nodes (superpixels) and absorbing nodes (duplicated nodes). The nodes in these two graphs constitute following three properties: (1) The nodes (including transient or absorbing) are associated with each other when superpixels in the image are adjacent nodes or have the same neighbors. And also boundary nodes (superpixels on the boundary of image) are fully connected with each other to reduce the geodesic distance between similar superpixels. (2) Any pair of absorbing nodes (which are duplicated from the boundaries or foreground nodes) are not connected (3) The nodes, which are duplicated from the four boundaries or foreground prior nodes, are also connected with original duplicated nodes. In this paper, the weight $w_{ij}$ of the edges is defined as
\begin{equation}
w_{ij}=e^{-\frac{\left \| x_{i}-x_{j} \right \|}{\sigma ^{^{2}}}}, i,j \in V^1 \ \text{or} \  i,j \in V^2
\end{equation}
where $\sigma$ is the constant parameter to adjust the strength of the weights in CIELAB color space. Then we can get the affinity matrix $A$
\begin{equation}
a_{ij}=
\begin{cases}
w_{ij}, &\text{ if $j\in M(i)$ \quad $1 \leq i \leq j$} \\
1, &\text{ if $i=j$ } \\
0, &\text{ otherwise},
\end{cases}
\end{equation}
where $M(i)$ is a nodes set, in which the nodes are all connected to nodes $i$. The diagonal matrix is given as: $D =  diag(\sum_{j}a_{ij})$, and the obtained transient matrix is calculated as:  $P = D^{-1} \times A.$

\subsection{Saliency detection model}
Following the aforementioned procedures, the initial image is transformed into superpixels, now two kinds of absorbing nodes for saliency detection are required. Firstly, we choose boundary nodes and foreground prior nodes to duplicate as absorbing nodes and obtain the absorbed times of transient nodes as foreground possibility and background possibility. Secondly, we use a cost function to optimize two possibility results together and obtain saliency results of all transient nodes.

\subsubsection{Absorb Markov chain via boundary nodes}
In normal conditions, four boundaries of an image rarely have salient objects. Therefore, boundary nodes are assumed as background, and four boundaries nodes set $H^1$ are duplicated as absorbing nodes set $D^1$, $H^1,D^1 \subset V^1$. The graph $G^1$ is constructed and absorbed time $z$ is calculated via Eq.\ref{Eq::absorb}. Finally, foreground possibility of transient nodes $z^f = \bar{z}(i) \quad i=1,2,\cdot\cdot\cdot,n,$ is obtained, and $\bar{z}$ denotes the normalizing the absorbed time vector.

\subsubsection{Absorb Markov chain via foreground prior nodes}
\begin{figure}[htb]
\centering
\includegraphics[width=1\textwidth]{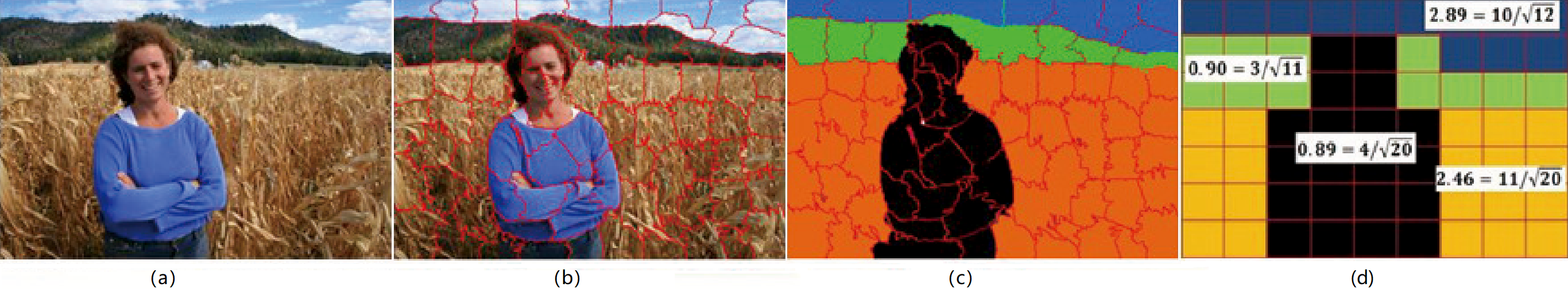}
\caption{An illustrative example of boundary connectivity. (a) input image (b)  the superpixels of input image (c) the superpixels of similarity in each pitches (d) an illustrative example of boundary connectivity.}
\label{fig::example}
\end{figure}

We use boundary connectivity to get the foreground prior $\textbf{f}_i$ without using the down-top method~\cite{zhu2014saliency}.
\begin{equation}\label{EQ:PF}
f_i = \sum^N_{j=1} (1 - \mathrm{exp}\big(-\frac{BC_j^2}{2\sigma_b^2}\big))d_a(i,j)\mathrm{exp}\big(-\frac{d_s^2(i,j)}{2\sigma_s^2}\big)
\end{equation}
where $d_a(i,j)$ and $d_s(i,j)$ denote the CIELAB color feature distance and spatial distance respectively between superpixel $i$ and $j$, the boundary connectivity (BC) of superpixel $i$ is defined as $BC_i = \frac{\sum_{j\in \mathcal{H}}w_{ij}}{\sqrt{\sum^N_{j=1}w_{ij}}}$ in Fig.~\ref{fig::example}, $\sigma_b = 1 $, $\sigma_s = 0.25 $. $\mathcal{H}$ denotes the boundary area of image and $w_{ij}$ is the similarity between nodes $i$ and $j$. $N$ is the number of superpixels. Afterwards, nodes ($\{i|f_i > avg(f)\}$) with high level values are selected to get a set $H^2$, which are duplicated as absorbing nodes set $D^2$, $H^2,D^2 \subset V^2$. The graph $G^2$ is constructed and absorbed time $z$ is calculated using Eq.\ref{Eq::absorb}. Finally, the background possibility of transient nodes $z^b = \bar{z}(i) \quad i=1,2,\cdot\cdot\cdot,n,$ is obtained, where $\bar{z}$ denotes the absorbed time vector normalization.

\subsection{Saliency Optimization}
In order to combine different cues, this paper has used the optimization model presented in \cite{zhu2014saliency}, which fused background possibility and foreground possibility for final saliency map.
It is defined as
\begin{equation}
\label{EQ::optimize}
\sum_{i=1}^{N}z^b_{i}s_{i}^{2}+\sum_{i=1}^{N}z^f_{i}(s_{i}-1)^{2}+\sum_{i,j}w_{ij}(s_{i}-s_{j})^{2}
\end{equation}
where the first term defines superpixel $i$ with large background probability $z^b$ to obtain a small value $s_i$ (close to 0).
The second term encourages a superpixel $i$ with large foreground probability $z^f$ to obtain a large value $s_i$ (close to 1).
The third term defines the smoothness to acquire continuous saliency values.

In this work, the used super-pixel numbers $N$ are 200, 250, 300 in the superpixel element, and  the final saliency map is given as: $\textbf{S} = \sum_h{S^h}$ at each scale, where $h = 1, 2, 3$.
The algorithm of our proposed method is summarized in Algorithm~\ref{alg::optimization}.
\begin{algorithm}[htb]
\caption{Saliency detection by bidirectional Markov chain.}
\label{alg::optimization}
\begin{algorithmic}[1]
\REQUIRE
An image and required parameters as the input image.
\STATE Using SLIC method to segment the input image into superpixels, construct two graphs $G^1(V^1,E^1)$ and $G^2(V^2,E^2)$;
\STATE Duplicate the four boundaries nodes and foreground prior nodes(which is computed by Eq.~\ref{EQ:PF}) as the absorbing nodes on graphs $G^1(V^1,E^1)$ and $G^2(V^2,E^2)$, respectively;
\STATE Compute the transient matrix $P$ with Eg.~\ref{Eq::transient time} on both graphs;
\STATE Obtain the absorbed time $z^f$ and $z^b$ with Eq.~\ref{Eq::absorb}, respectively;
\STATE Optimize the model with the Eq.~\ref{EQ::optimize};
\STATE Compute final saliency values with $\textbf{S} = \sum_{h=1}^{3}\textbf{S}^h$;
\ENSURE
Output a saliency map $\textbf{S}$ with the same size as the input image.
\end{algorithmic}
\end{algorithm}

\section{Experiments}
The proposed method is evaluated on four benchmark datasets ASD~\cite{achanta2009frequency}, CSSD~\cite{yan2013hierarchical}, ECSSD~\cite{yan2013hierarchical} and SED~\cite{alpert2012image}.
ASD dataset is a subset of the MSRA dataset, which contains 1000 images with accurate human-labeled ground truth.
CSSD dataset, namely complex scene saliency detection contains 200 complex images.
ECSSD dataset, an extension of CSSD dataset contains 1000 images and has accurate human-labeled ground truth.
SED dataset has two parts, SED1 and SED2, images in SED1 contains one object, and images in SED2 contains two objects, in total they contain 200 images.
We compare our model with 17 different state-of-the-art saliency detection algorithms: CA~\cite{goferman2012context}, FT~\cite{achanta2009frequency},  SEG~\cite{rahtu2010segmenting}, BM~\cite{xie2011visual}, SWD~\cite{duan2011visual}, 
SF~\cite{perazzi2012saliency}, GCHC~\cite{yang2013graph}, LMLC~\cite{xie2013bayesian}, HS~\cite{yan2013hierarchical}, PCA~\cite{margolin2013makes}, DSR~\cite{li2013saliency}, MC~\cite{jiang2013saliency}, MR~\cite{yang2013saliency}, MS~\cite{tong2014saliency}, RBD~\cite{zhu2014saliency}, RR~\cite{li2015robust}, MST~\cite{tu2016real}. The tuning parameters in the proposed algorithm is the edge weight $\sigma^2=0.1$ that controls the strength of weight between a pair of nodes. In the following results of the experiments, we show the evaluation of our proposed saliency models based on the aforementioned datasets comparing with the best works. In addition, we also give some limitation about our model and analysis the reason.

\subsection{Evaluation of the proposed model}
\begin{figure*}[htb]
  \centering
   \includegraphics[scale=0.28]{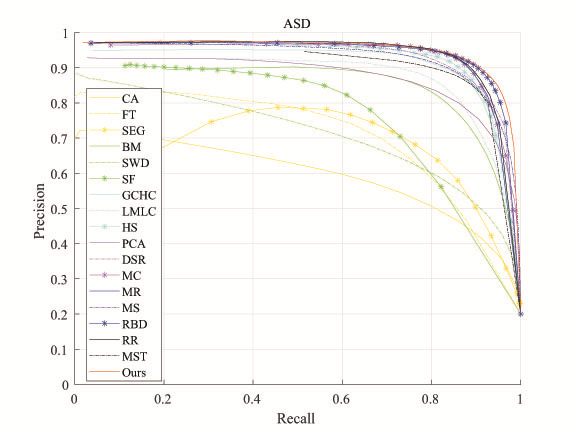}
   \includegraphics[scale=0.38]{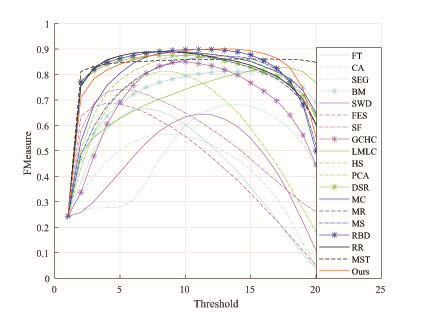}
   \caption{PR-curves and F-measure curves comparing with different methods on ASD dataset.}
\normalsize
\label{fig::PR1}
\end{figure*}

\begin{figure*}[h]
  \centering
   \includegraphics[scale=0.28]{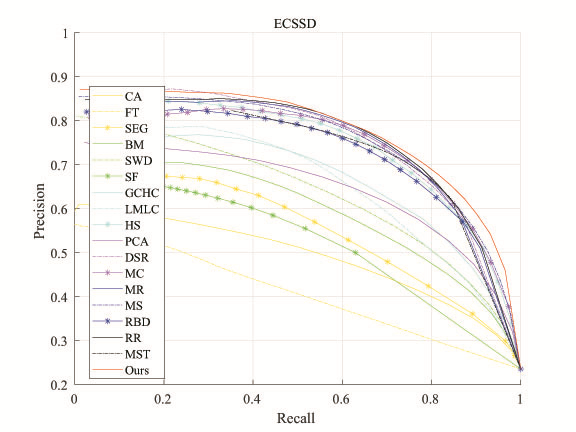}
   \includegraphics[scale=0.38]{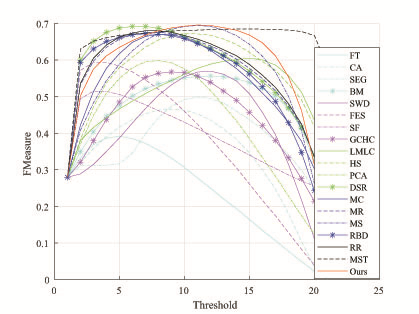}
   \caption{PR-curves and F-value curves comparing with different methods on ECSSD dataset.}
\normalsize
\label{fig::PR2}
\end{figure*}

\begin{figure*}[htb]
  \centering
   \includegraphics[scale=0.28]{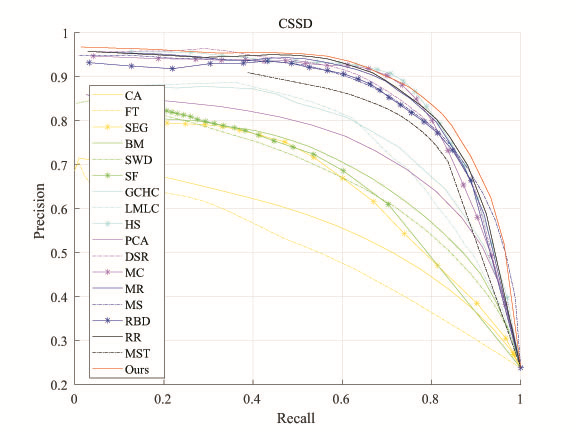}
   \includegraphics[scale=0.38]{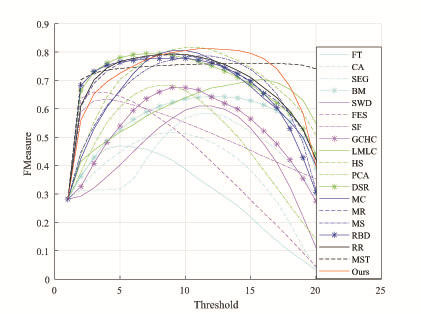}
   \caption{PR-curves and F-measure curves comparing with different methods on CSSD dataset.}
\normalsize
\label{fig::PR3}
\end{figure*}

\begin{figure*}[htb]
  \centering
   \includegraphics[scale=0.28]{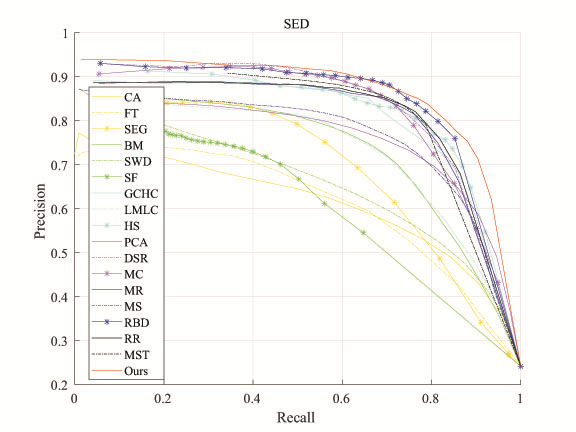}
   \includegraphics[scale=0.38]{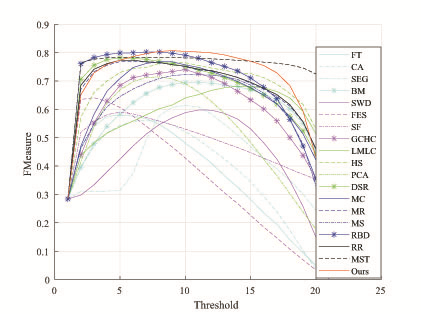}
   \caption{PR-curves and F-measure curves comparing with different methods on SED dataset.}
\normalsize
\label{fig::PR4}
\end{figure*}

\begin{figure}[htb]
  \centering
   \includegraphics[width=1\textwidth]{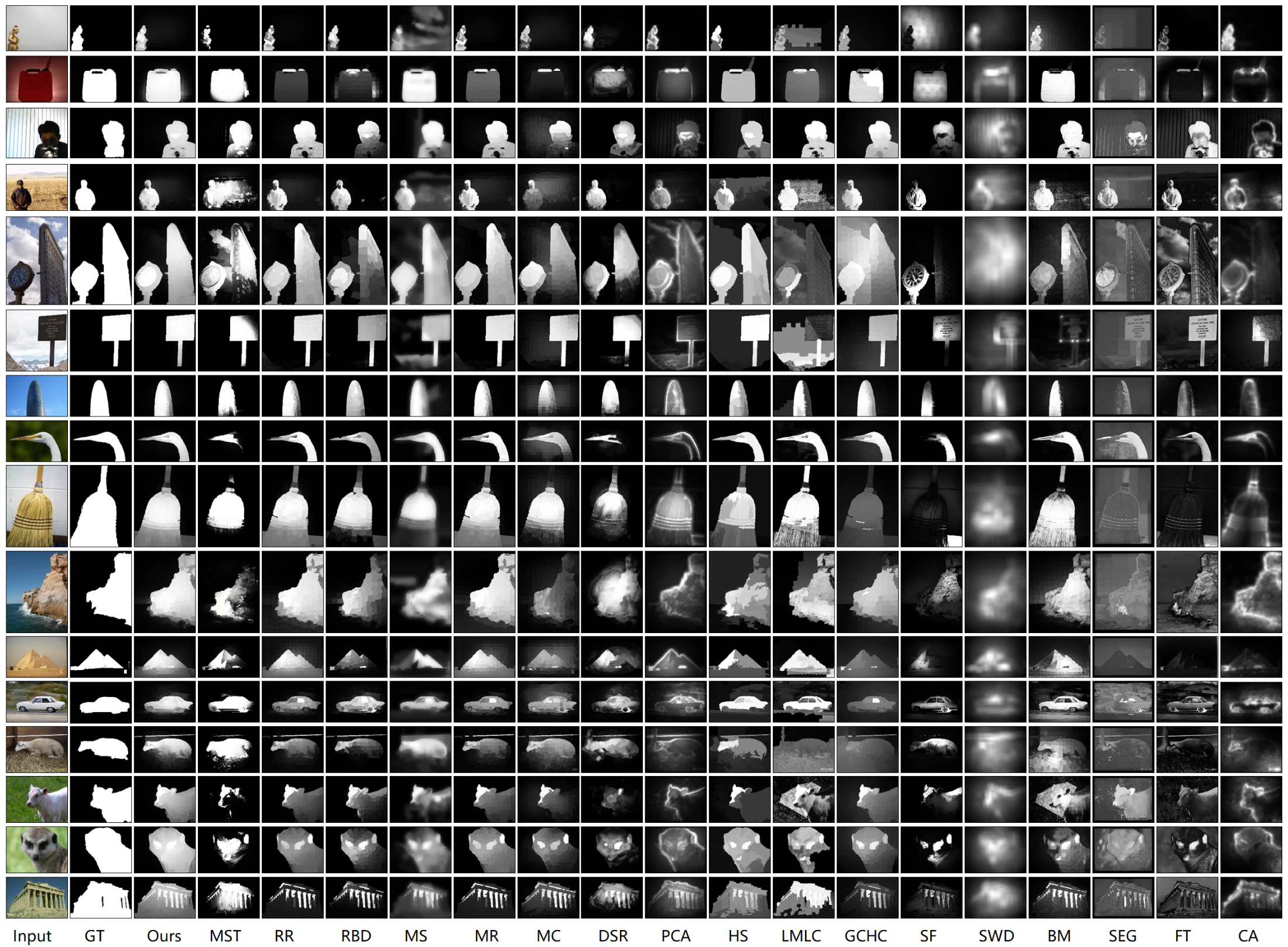}
   \caption{Examples of output saliency maps results using different algorithms on the ASD, CSSD, ECSSD and SED datasets}
\normalsize
\label{fig::display}
\end{figure}

The precision-recall curves and F-measure are used as performance metrics. The precision is defined as the ratio of salient pixels correctly assigned to all the pixels of extracted regions. The recall is defined as the ratio of detected salient pixels to the ground-truth number. Which can be fomulated as,
\begin{equation}
Precision = \frac{TP}{TP+FP},Recall = \frac{TP}{TP+FN}£¬
\end{equation}
where $TP$, $FP$ and $FN$ represent the true positive, false positive and false negative, respectively. A PR curve is obtained by the threshold sliding from 0 to 255 to get the difference between the saliency map (which is calculated) and ground truth(which is labeled manually). F-measure is calculated by the weighted average between the precision values and recall values, which can be regarded as overall performance measurement, given as:
\begin{equation}
F_{\beta } = \frac{(1+\beta ^{2})Precision \times Recall}{\beta ^{2}Precision + Recall},
\end{equation}
we set $\beta^{2} = 0.3$ to stress precision more than recall.
PR-curves and the F-measure curves are shown in Figure~\ref{fig::PR1} - \ref{fig::PR4}, where the outperformance of our proposed method as compared to 17 state-of-the-art methods is evident. Fig.\ref{fig::display} presets visual comparisons selected from four datasets. It can be seen that the proposed method achieved best saliency results as compared to the state-of-the-art methods.

\subsection{Failure cases analysis}
\begin{figure}[htb]
  \centering
   \includegraphics[width=0.7\textwidth]{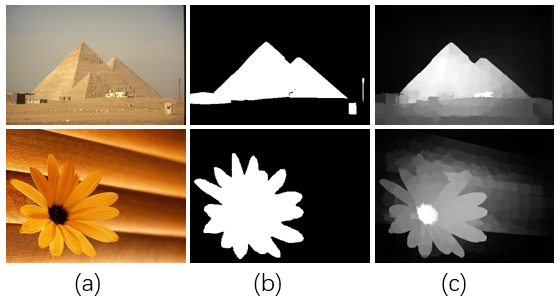}
   \caption{Examples of our failure examples. (a) Input images. (b) Ground truth. (c) Saliency maps.}
\normalsize
\label{fig::failure}
\end{figure}

In this work, the idea of bidirectional absorbing Markov chain is first proposed. Although the proposed method is effective for most images on the four datasets. However, if the appearances of four boundaries and the foreground prior are similar to each other, the performance is not obviously, which is shown in Figure~\ref{fig::failure}.

\section{Conclusion}
In this paper, a bidirectional absorbing Markov chain based saliency detection method is proposed considering both boundary information and foreground prior cues. A novel optimization model is developed to combine both background and foreground possibilities, acquired through bidirectional absorbing Markov chain. The proposed approach outperformed 17 different state-of-the-art methods over four benchmark datasets, which demonstrate the superiority of our proposed approach. In future, we intend to apply our proposed saliency detection algorithm to problems such as multi-pose lipreading and audio-visual speech recognition.

\section{Acknowledgments}
This work was supported by China Scholarship Council, the National Natural Science Foundation of China (No.913203002), the Pilot Project of Chinese Academy of Sciences (No.XDA08040109). Prof. Amir Hussain and Dr. Ahsan Adeel were supported by the UK Engineering and Physical Sciences Research Council (EPSRC) grant No.EP/M026981/1.




 \section{Reference}
 \bibliographystyle{elsarticle-num}
 \bibliography{BICS67}




\end{document}